%% file: main.tex
\documentclass[conference]{IEEEtran}
\IEEEoverridecommandlockouts
\usepackage{cite}
\usepackage{url}
\usepackage{amsmath,amssymb,amsfonts}
\usepackage{algorithmic}
\usepackage{graphicx}
\usepackage{textcomp}
\usepackage{xcolor}
\usepackage{url}
\def\BibTeX{{\rm B\kern-.05em{\sc i\kern-.025em b}\kern-.08em
    T\kern-.1667em\lower.7ex\hbox{E}\kern-.125emX}}
    
\usepackage{tikz}
\usepackage{textcomp}
\usepackage[doipre={DOI:~}]{uri}
\usepackage{lipsum}
\newcommand\copyrighttext{%
  \footnotesize \textcopyright 2022 IEEE. Personal use of this material is permitted. Permission from IEEE must be obtained for all other uses, in any current or future media, including reprinting/republishing this material for advertising or promotional purposes, creating new collective works, for resale or redistribution to servers or lists, or reuse of any copyrighted component of this work in other works.
 
  Accepted as a conference paper at the 2022 IEEE International Conference on Ph. D. Research in Microelectronics and Electronics (PRIME).}
  
\newcommand{\copyrightnotice}{%
\begin{tikzpicture}[remember picture,overlay]
\node[anchor=south,yshift=10pt] at (current page.south) {\fbox{\parbox{\dimexpr\textwidth-\fboxsep-\fboxrule\relax}{\copyrighttext}}};
\end{tikzpicture}%
}

\begin{document}

\title{Energy-efficient and Privacy-aware Social Distance Monitoring with Low-resolution Infrared Sensors and Adaptive Inference
\thanks{This work has received funding from the ECSEL Joint Undertaking (JU) under grant agreement No 101007321. The JU receives support from the European Union’s Horizon 2020 research and innovation programme and France, Belgium, Czech Republic, Germany, Italy, Sweden, Switzerland, Turkey.}
}

\author{
\IEEEauthorblockN{Chen Xie}
\IEEEauthorblockA{
\textit{DAUIN, Politecnico di Torino}\\
Turin, Italy \\
chen.xie@polito.it}
\and
\IEEEauthorblockN{Daniele Jahier Pagliari}
\IEEEauthorblockA{
\textit{DAUIN, Politecnico di Torino}\\
Turin, Italy \\
daniele.jahier@polito.it}
\and
\IEEEauthorblockN{Andrea Calimera}
\IEEEauthorblockA{
\textit{DAUIN, Politecnico di Torino}\\
Turin, Italy \\
andrea.calimera@polito.it}
}

\maketitle
\copyrightnotice

\begin{abstract}
\input{sec/abstract}

\end{abstract}


\section{Introduction and Related Works}

\input{sec/introduction}

\section{Proposed Method}\label{sec:method}

\input{sec/method}

\section{Experimental Results}

\input{sec/results}

\section{Conclusions}
\input{sec/conclusion}

\bibliographystyle{IEEEtran}

\end{document}

%% file: sec/abstract.tex
Low-resolution infrared (IR) Sensors combined with machine learning (ML) can be leveraged to implement privacy-preserving social distance monitoring solutions in indoor spaces. However, the need of executing these applications on Internet of Things (IoT) edge nodes makes energy consumption critical. In this work, we propose an energy-efficient adaptive inference solution consisting of the cascade of a simple wake-up trigger and a 8-bit quantized Convolutional Neural Network (CNN), which is only invoked for difficult-to-classify frames.
Deploying such adaptive system on a IoT Microcontroller,  we show that, when processing the output of a 8x8 low-resolution IR sensor, we are able to reduce the energy consumption by 37-57\% with respect to a static CNN-based approach, with an accuracy drop of less than 2\% (83\% balanced accuracy).

%% file: sec/introduction.tex
As one of the most effective ways to avoid catching an infectious disease, particularly in densely populated areas, social distancing~\cite{sun2020efficacy} has demonstrated its usefulness to combat the spread of COVID-19. In this scenario, the demand for automated social distance monitoring solutions has increased dramatically, especially for public indoor environments such as shops, offices, etc.

Researchers have proposed multiple technical solutions for this task. A first approach consists of computing the distance between people using the transceivers embedded in their personal wearable devices or smartphones~\cite{li2021smartdistance}. However, since this solution is heavily dependent on the voluntary participation by users, it is not easy to guarantee its effectiveness in a real-world case. Another method relies on IoT cameras that monitor a specific area, track individuals, and compute social distance with Machine Learning (ML) or Deep Learning (DL) algorithms, executed either at the edge or in the cloud~\cite{yang2021vision}. While this approach eliminates the requirement of active user participation,
it creates new issues related to privacy. In fact, it permits not only to spot social distance violations, but also to identify and track individuals, often in violation with privacy protection laws. Furthermore, leaks of private information (either accidental or caused by malicious parties) are possible.

Based on these observations, low-resolution infrared (IR) array sensors configure as a promising alternative~\cite{xie2022privacy}. In fact, these sensors are composed by a small number of thermal pixels (8x8 or 16x16), which only capture the basic shapes and temperatures of objects, without having a sufficient resolution to reveal private information. Moreover, they are effective regardless of the lighting conditions, including at night. Last, and most importantly, with their low power consumption and low-cost, they are fully compatible with an edge implementation. Combined with ML or DL algorithms, they allow the realization of a \textit{real-time} monitoring system without any permanent data storage or transmission to the cloud, which further contribute to privacy protection.

In recent years, many researchers have studied applications of ML to low-resolution IR sensors data~\cite{mashiyama2015activity,shih2020multiple,herrmann2018cnn, trofimova2017indoor,Metwaly2019,xie2022privacy}. However, these studies focused on different tasks, such as human activity recognition~\cite{mashiyama2015activity, shih2020multiple}, presence detection~\cite{trofimova2017indoor, herrmann2018cnn} or people counting~\cite{Metwaly2019}.
To our knowledge, our previous work of \cite{xie2022privacy} is the only dedicated implementation of a social distance monitoring system on a low-resolution IR sensor, based on Convolutional Neural Networks (CNNs).

Despite the aforementioned advantages of IR array sensors, the realization of a ML/DL-based system for social distance monitoring at the edge remains critical from the point of view of energy consumption. In fact, IoT end-nodes are typically battery-operated, and based on general-purpose Microcontrollers (MCUs), since their tight cost budgets do not allow the luxury of hardware specialization to improve energy efficiency.
In our previous work, we optimized the consumption of our CNN using 8-bit quantization~\cite{Jacob2018}, a well known technique that reduces the numeric format used to represent the inputs, intermediate activations, and parameters of the model.
While effective, however, quantization is a \textit{static} optimization, applied identically to all inputs. As such, it misses opportunities for energy saving, as it cannot exploit the fact that not all inputs are equally hard to classify. Intuitively, a CNN may be useful to distinguish whether the ``heat shape'' measured by the IR sensor corresponds to a single person (OK) or two nearby people (social distance violation). In contrast, a much simpler classifier is sufficient to just detect if \textit{at least one person} is present in the frame.

In this work, we start from this observation and propose the first \textit{adaptive inference} system for low-resolution IR arrays. The term adaptive inference refers to a \textit{run-time} optimization, in which the complexity of the computation is automatically tuned to the difficulty of the processed input. Examples of adaptive systems include big/little models~\cite{Park2015,JahierPagliari2018a}, early-exit~\cite{branchy,daghero2021adaptive}
and hierarchical/staged classifiers~\cite{hierarchical}.
We follow the latter approach, proposing a simple yet effective hierarchical classifier based on a lightweight wake-up trigger, that triggers the CNN execution \textit{only for inputs that require it}. 
Using the same dataset and CNN classifier of~\cite{xie2022privacy}, and deploying our models on Quentin~\cite{schiavone2018quentin}, a 22-nm 32bit single-core RISC-V MCU, we show that we can reduce the total energy consumption by 37-57\% with respect to our previous work, with a limited accuracy drop $<$ 2\%.

%% file: sec/method.tex
\subsection{Dataset and Social Distance Monitoring Solution}

Under proper conditions, social distancing can be framed as a variant of people counting.
In fact, based on the IR sensor viewing range, it is possible to position the monitoring device such that a social distance violation occurs \textit{whenever more than a given number of people are in the frame}.
This, in turn, reduces the problem to a binary classification (OK/violation).

\begin{figure}[ht]
\centering
\includegraphics[width=.7\columnwidth]{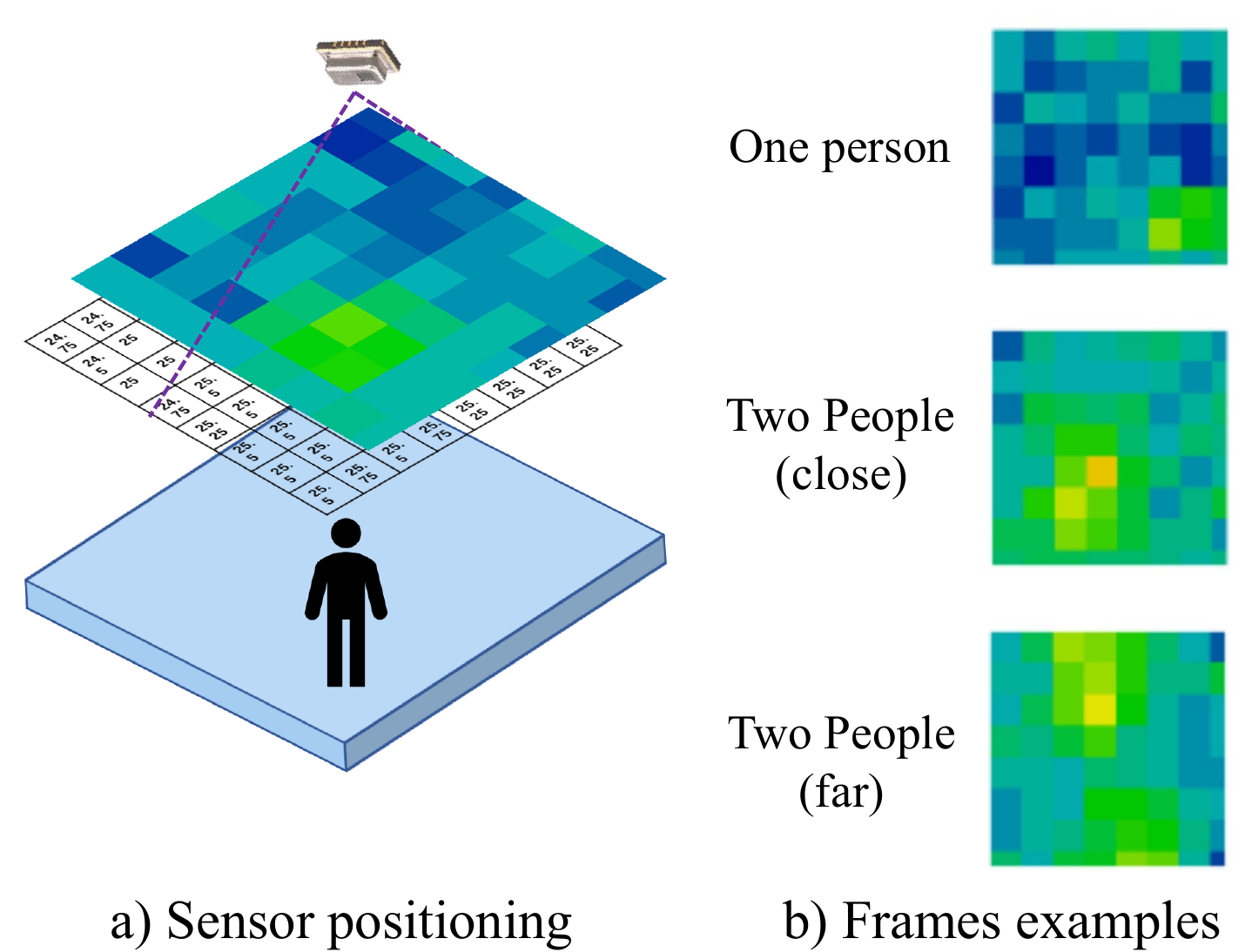}
\caption{Grid-EYE sensor positioning and IR frames examples.}\label{fig:sensor}
\end{figure}

In this work, we use the open-source \textit{LINAIGE} dataset~\cite{xie2022privacy}, which is specifically tailored for person counting and presence detection tasks. It contains low-resolution 8x8-pixel infrared sensor samples, associated with the corresponding people count. The samples are collected in different indoor environments at 10 Frames Per Second (FPS), by a ceiling-mounted Panasonic Grid-EYE AMG8833 sensor~\cite{grideyeapi} , with a view angle of 60° (see Fig.~\ref{fig:sensor}a).
The approximate view range of the IR sensor can be calculated from the view angle and the distance between sensor and detected objects (people heads), as pointed out in~\cite{grideyeapi,xie2022privacy}. For LINAIGE data, the maximum distance within one frame (the length of the diagonal in the squared area) is in the [1.53:2.04] m range. Combining this with the typical 2m social distancing recommendation~\cite{voko2020effect}, we can conclude that a social distance violation is present in the view area of the sensor \textit{whenever 2 or more people appear in a single frame}. Larger areas can be covered by combining and coordinating multiple IR sensors (out of scope in this work).

This problem formulation leads naturally to the use of an adaptive inference system. In fact, the problem can be decomposed in two parts. First, we should distinguish between an empty frame and people presence; this is a simple task, that can be solved by a very light model. Only when people presence is spotted, we then have to distinguish between 1 or more people; this is more complex, since the heat shapes produced by nearby people are often overlapped, especially with such low-resolution sensors, as shown in Fig.~\ref{fig:sensor}b. Therefore, a model of higher complexity is required. However, since in a real-world scenario it is likely that many frames will be empty, invoking the more complex model only when presence is spotted could significantly reduce the average energy consumption of the system.

\begin{figure*}[ht]
\centering
\includegraphics[width=\textwidth]{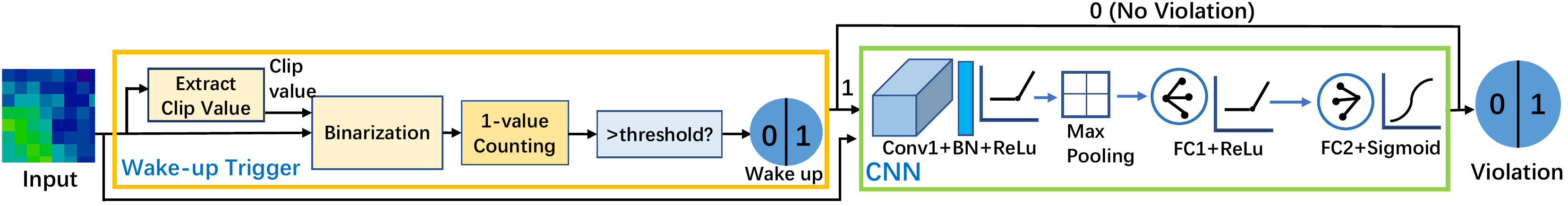} 
\caption{Overview of the proposed adaptive inference system.} \label{fig:adaptive}
\end{figure*}

\subsection{Wake-up Trigger and CNN}

In order to implement the proposed adaptive inference system, we combine a simple and deterministic classifier, which we call \textit{Wake-up Trigger}, with a compact and 8-bit quantized CNN.
The whole adaptive framework is shown in Fig.~\ref{fig:adaptive}. 
As anticipated, we split the social distance monitoring task into two stages, each formulated as a binary classification. In the first stage, the wake-up trigger detects whether at least one person is present in the frame. If this first stage results in a ``no-person'' prediction, inference is terminated immediately with a 0 output (no violation of social distancing). Otherwise, the second stage is triggered, and another binary classification is performed using the CNN, to produce the final violation/no-violation output.

As shown in Fig.~\ref{fig:adaptive}, the Wake-up Trigger is substantially a threshold-based classifier.
It works by binarizing the IR frame pixels to 0/1 based on the measured temperature. Then, the number of 1-valued pixels is counted and compared with a threshold. If the threshold is exceeded, it is assumed that at least one person is present in the frame, and the CNN is invoked. The clipping value used for binarization is computed periodically by analyzing a window of N consecutive frames with no people detected, and extracting the \textit{maximum} temperature in all pixels (\textit{Extract Clip Value} block in the figure). The rationale is that the head of a person passing under the IR array will be sensed as a \textit{temperature increase} with respect to the background, also due to the lower distance from the measuring device with respect to background heat sources (e.g., heaters).
Indeed, the input binarization implemented by the Wake-up Trigger can be seen as a sort of background removal process. %
In our experiments, we set $N=8$.

When human presence is detected by the Wake-up trigger, the CNN classifier is invoked. We select this type of DL model due to the state-of-the-art results achieved on multiple visual recognition tasks, and we use a 8-bit quantized CNN architecture similar to the one originally proposed in~\cite{xie2022privacy} to predict social distancing violations. 
In particular, among the multiple CNN variants proposed in that paper, we select the one achieving the \textit{best balanced accuracy} performance, whose structure is shown in Fig.~\ref{fig:adaptive}. The only difference with respect to the original architecture is the addition of Batch Normalization (BN), which substantially improves the final classification accuracy.
The CNN consists of one Convolutional (Conv) layer with BN and Rectified Linear Unit (ReLU) activation, one Max Pooling layer and two fully-connected (FC) layers with ReLU and sigmoid activations respectively. The number of channel in Conv layer is set to 64 and the first FC layer has a hidden size of 64 as well.

%% file: sec/results.tex
We trained our CNN model using Keras on the \textit{LINAIGE}~\cite{xie2022privacy} multi-pixel IR dataset using a per-session train/test split. Namely, the data collected in Session 1 have been used as training set, while all other sessions as the test set. This ensures that test data are collected in a different environment and/or at a different day/time compared to training data. Each training has been repeated with 5 different random seeds. We trained for a maximum of 500 epochs with a 10-epoch patience early stopping. The learning rate was set to $10^{-3}$ and reduced by a factor 0.3 after 5 non-improving epochs. The CNNs have been quantized to 8-bit with the built-in quantization-aware training of Keras. 
We measured classification quality using the standard accuracy, also the \textit{balanced accuracy} (Bal. Acc.) and the F1-Score, both of which are more meaningful for highly class-imbalanced datasets. For results involving the CNN, we report mean $\pm$ standard deviation over the 5 training runs.

We deployed all models on Quentin~\cite{schiavone2018quentin}, a 22nm, 32bit single-core RISC-V platform with 520kB of memory.
Latency and energy consumption for inference are estimated using the GVSoC virtual platform~\cite{bruschi2021gvsoc}, for a clock frequency of 205.1 MHz and a supply voltage of 0.54V. All deployed models are written in C; for CNNs, we used the optimized kernels library presented in~\cite{pulp_nn}, in its single-core version.

\subsection{Classification Performance and Energy Savings}

\begin{figure*}[ht]
\centering
\includegraphics[width=.94\linewidth]{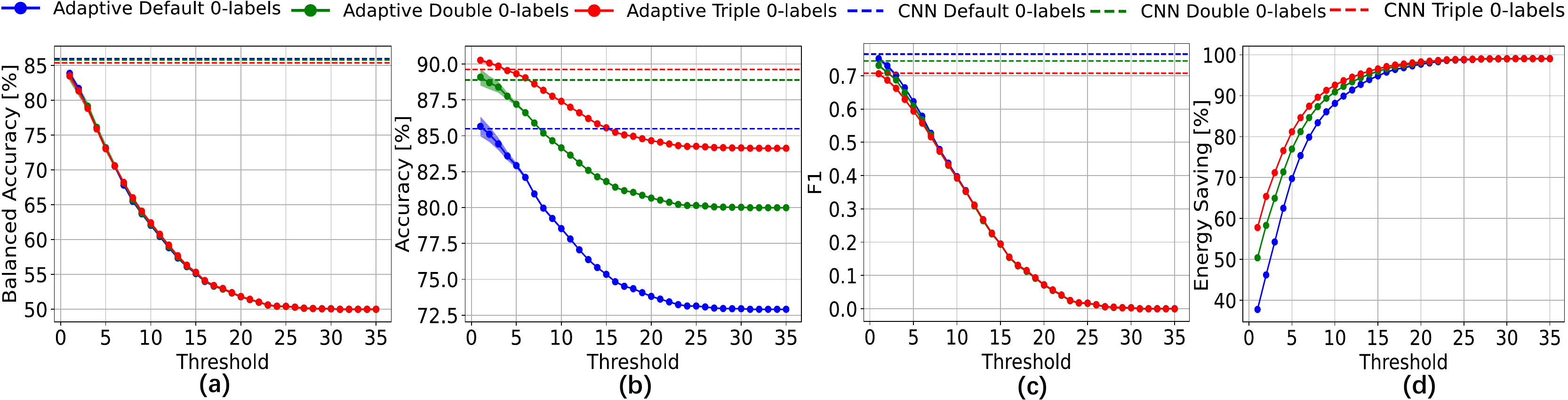}
\caption{Adaptive System Performance in terms of Threshold vs Balanced Accuracy (a), vs Accuracy (b), vs F1 (c) and vs Energy Saving (d).}\label{fig:curves}
\vspace{-0.4cm}
\end{figure*}

\begin{table*}[ht]
    \centering
     \caption{Detailed evaluation and deployment Results of Models Targeting Social Distancing on Quentin @ 205.1 MHz \label{tab:model}}
    \begin{tabular}{|c|c|c|c|c|c|c|c|}
    \hline
      \textbf{Model} & \textbf{Bal. Acc. [\%]}  & \textbf{Acc. [\%]}
      & \textbf{F1} & \textbf{Model Mem. [B]} & \textbf{Tot. Mem. [B]}  &  \textbf{Energy[µJ]} & \textbf{Latency[µs]}  \\  \hline
      8-bit CNN~\cite{xie2022privacy} & 85.96±0.47  & 85.50±1.05 & 0.76±0.01  & 37.97k & 56.07k  & 1.20  & 316 \\ \hline
    
     Wake-up Trigger & 71.30 & 65.28 & 0.57 & 66 & 12.25k & 0.01 & 2.96  \\ \hline
    
    \cite{grideye} & 61 & 73 & 0.41 & 1.71k &  23.31k & 2.53 & 665 \\ \hline
    \textbf{Our Method} & 83.86±0.29 & 85.65±0.68 & 0.75±0.01 & 38.04k & 56.24k & 0.75 & 198 \\
    
    \hline
     \end{tabular}
\end{table*}

The solid lines and dots in Fig.~\ref{fig:curves}a-c show the classification performance obtained by the proposed adaptive system in terms of balanced accuracy, accuracy, and F1-score respectively. Different points are obtained varying the configurable threshold of the wake-up trigger, and correspond to the mean score obtained over the 5 training runs.  Standard deviation is reported as a colored band (hardly visible in most curves).
Dashed lines refer to the results obtained by a static CNN identical to the second ``stage'' of our adaptive method. Lastly, Fig.~\ref{fig:curves}d shows the average energy savings in percentage obtained by the adaptive system, compared to the static CNN.

The three colors correspond to three different versions of the LINAIGE test set.
Specifically, following the observation of Sec.~\ref{sec:method} that in a practical application scenario, it is likely to observe long sequences of frames with no people in them, we consider three variants of the test data: i) \textit{Default}, the original LINAIGE test set, ii) \textit{Double}, a version of the test set with all frames corresponding to a ground truth label equal to 0 people duplicated, and iii) \textit{Triple}, similar to the previous one, but with triplicated 0-labels.

A first result shown by the figure is that the overall performance of the adaptive system decreases for larger threshold values, and that the best results are obtained for threshold $=1$. This is expected, since a higher threshold corresponds to fewer invocations of the second-stage CNN. For threshold values $>25$, the adaptive system stops working completely, and begins to act as a constant classifier (always predicting 0).
Vice versa, lower thresholds are more conservative: they reduce the number of false negatives, at the cost of potentially more CNN invocations than needed, even for frames that do not actually contain people (i.e., false positives). Importantly, however, the CNN can still ``correct'' these false positives by predicting the ``No Violation'' class, thus improving the overall Accuracy, Bal. Acc and F1.
For the same reasons, the energy savings show an opposite trend, increasing together with the wake-up threshold (Fig.~\ref{fig:curves}d).

The most interesting results are obtained with a threshold of 1: the adaptive system reaches a balanced accuracy of around 83\%, only 2\% lower than a static CNN solution. Furthermore, the F1-Score and standard accuracy remain substantially identical to the static case. At the same time, by invoking the CNN only when presence is detected in the frame, the average energy consumption is reduced by 37\% to 57\% depending on the number of 0-labels. As expected, the savings increase when there are more frames without people, showing the potential effectiveness of the proposed light-weight wake-up trigger in a real-world scenario.

\subsection{Detailed Comparison}

Table~\ref{tab:model} compares the proposed adaptive system (with threshold $=$ 1) to the two individual classifiers (wake-up trigger and CNN), and to the state-of-the-art deterministic algorithm proposed in~\cite{grideye}. Besides the classification scores, the table also reports the data memory occupied by each model once deployed on the target MCU, the total memory including code size, and the average energy consumption and latency for a classification. The ``Wake-up Trigger'' row refers to the threshold-based classifier used ``stand-alone'', directly for social distance monitoring. In this case, the best performance are achieved with a threshold $=$ 2. All results refer to the \textit{Default} LINAIGE test set.

As shown, our method significantly outperforms both the stand-alone wake-up trigger and the approach of \cite{grideye} on all classification metrics (up to +20\% Bal. Acc.), while obtaining comparable results to the static CNN, which is analogous to the one presented in~\cite{xie2022privacy}. At the same time, the average energy and latency are significantly reduced w.r.t. the CNN (37.5\%), at the cost of a negligible total memory overhead ($<$0.3\%), demonstrating that the proposed adaptive system can be effectively deployed on MCU-class platforms.

%% file: sec/conclusion.tex
We have proposed an energy-efficient and privacy-aware social distance monitoring solution based on low-resolution IR arrays and adaptive inference. Our results on a low-power MCU show that energy saving over 37\% can be achieved with respect to a static CNN-based approach. Future works will include the exploration of adaptive inference combined with other DL models.